\documentclass[sigconf,review=false,anonymous=false,authordraft=false]{acmart}

\usepackage{arydshln}

\makeatother

\usepackage{enumitem}
\newlist{inlinelist}{enumerate*}{1}
\setlist*[inlinelist,1]{%
  label=(\roman*),
}
\usepackage{subfigure}
\usepackage{xspace}

\usepackage{amsmath}
\usepackage{multicol}
\usepackage{multirow}

\newcommand{\ourframework}{\textsc{uRAG}\xspace} 

\title{Towards a Search Engine for Machines: Unified Ranking for Multiple Retrieval-Augmented Large Language Models}

\author{Alireza Salemi}
\affiliation{\institution{University of Massachusetts Amherst}
\city{Amherst}
\state{MA}
\country{United States}}
\email{asalemi@cs.umass.edu}

\author{Hamed Zamani}
\affiliation{\institution{University of Massachusetts Amherst}
\city{Amherst}
\state{MA}
\country{United States}}
\email{zamani@cs.umass.edu}

\copyrightyear{2024}
\acmYear{2024}
\setcopyright{rightsretained}
\acmConference[SIGIR '24]{Proceedings of the 47th International ACM SIGIR Conference on Research and Development in Information Retrieval}{July 14--18, 2024}{Washington, DC, USA}
\acmBooktitle{Proceedings of the 47th International ACM SIGIR Conference on Research and Development in Information Retrieval (SIGIR '24), July 14--18, 2024, Washington, DC, USA}
\acmDOI{10.1145/3626772.3657733}
\acmISBN{979-8-4007-0431-4/24/07}

\begin{document}


\begin{abstract}
This paper introduces \ourframework--a framework with a unified retrieval engine that serves multiple downstream retrieval-augmented generation (RAG) systems. Each RAG system consumes the retrieval results for a unique purpose, such as open-domain question answering, fact verification, entity linking, and relation extraction. We introduce a generic training guideline that standardizes the communication between the search engine and the downstream RAG systems that engage in optimizing the retrieval model. This lays the groundwork for us to build a large-scale experimentation ecosystem consisting of 18 RAG systems that engage in training and 18 unknown RAG systems that use the \ourframework as the new users of the search engine. Using this experimentation ecosystem, we answer a number of fundamental research questions that improve our understanding of promises and challenges in developing search engines for machines. 
\end{abstract}

\keywords{Retrieval-enhanced machine learning; retrieval augmentation; neural ranking model; large language model; text generation}

\ccsdesc[500]{Computing methodologies~Natural language generation}
\ccsdesc[500]{Information systems~Information retrieval}
\ccsdesc[500]{Information systems~Retrieval models and ranking}
\ccsdesc[500]{Information systems~Question answering}
\ccsdesc[500]{Computing methodologies~Machine learning}

\maketitle

\section{Introduction}
\label{sec:intro}
The vast majority of machine learning systems, including large generative models, are designed as self-contained systems,  with both knowledge and reasoning encoded in model parameters. 
However, these models cannot work effectively for tasks that require knowledge grounding \cite{DBLP:journals/corr/abs-2305-15771,openai-retrieval-plugin}, especially in case of non-stationary data where new information is actively being produced \cite{reml,freshqa23}. 
As suggested by \citet{reml}, this issue can be addressed when machine learning systems are being \emph{enhanced with the capability of retrieving stored content}. For example, in retrieval-augmented generation (RAG), as a special case of retrieval-enhanced machine learning (REML) \cite{reml}, systems consume the responses provided by a retrieval model for the purpose of text generation \cite{rag, li2022survey}. 
RAG models demonstrate substantial promise across various applications, including open-domain question answering \cite{siriwardhana-etal-2023-improving, rag, zhu2021retrieving, dpr, cheng-etal-2021-unitedqa}, fact verification \cite{thorne-etal-2018-fever}, dialogue systems \cite{weston-etal-2018-retrieve, dinan2018wizard, ijcai2018p609}, machine translation \cite{cai-etal-2021-neural}, and personalized generation \cite{salemi2024lamp, salemi2024optimization}.

In the RAG literature, the retrieval component is often implemented using either of the following two approaches:
\begin{enumerate}[leftmargin=*]
    \item Employing an off-the-shelf retrieval model that does not require training for the downstream RAG system: in this category, RAG systems either use APIs from commercial web search engines \cite{nakano2022webgpt}, term matching retrieval models \cite{glass-etal-2022-re2g}, such as TF-IDF and BM25, or neural ranking models trained on relevance annotations provided as an external resource \cite{rag};

    \item Training a retrieval model given the feedback from the downstream RAG system through knowledge distillation \cite{distil-fid} or end-to-end optimization \cite{sachan-etal-2021-end}.
\end{enumerate}

\begin{figure}
    \centering
    \includegraphics[width=\linewidth]{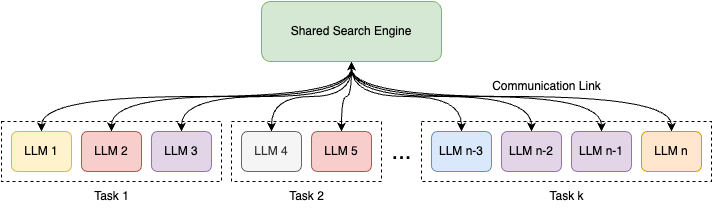}
    \caption{A high-level overview of the \ourframework ecosystem. The ecosystem consists of a shared search engine that serves multiple RAG models, each performing its own task.}
    \label{fig:overview-problem}
\end{figure}

As expected, the latter category offers the current state-of-the-art performance for various tasks \cite{rl-retriever-train, atlas}. From an IR perspective, in this category, the downstream RAG model is the only ``user'' of the search engine. This paper takes a step further by relying on an important lesson learned by the IR community since the creation of web search: \emph{achieving an effective retrieval model can be achieved through aggregating large-scale implicit feedback (e.g., clicks) obtained across users} \cite{10.1145/1076034.1076045}. Following this lesson, we aim to develop \ourframework, a framework for training a single \underline{u}nified search engine for multiple \underline{RAG} systems conducting various downstream tasks (see Figure~\ref{fig:overview-problem}). We define a training guideline, in which the search engine and its users---multiple downstream RAG systems---engage in an optimization process, in which RAG systems submit queries and the search engine solicits feedback for a number of returned documents in their response. Each downstream RAG model identifies a utility function (any arbitrary evaluation metric; differentiable or not) and uses it for providing scalar feedback to the search engine. 

Using such a generic optimization guideline by \ourframework, we implement a large-scale experimentation ecosystem that enables addressing important research questions in this area. In more detail, we implement a set of 18 RAG systems that use different large language model (LLM) architectures, conduct different tasks, consume different number of retrieved documents, and/or trained on different datasets. We consider three diverse tasks of open-domain question answering, fact verification, and slot-filling for relation extraction. Six datasets related to these tasks from the knowledge-intensive language tasks (KILT) benchmark \cite{kilt} are used for training and evaluation.  In addition to these 18 RAG models that engage with our unified search engine for optimization, we consider another set of 18 RAG models with distinct properties as the \emph{new users} of \ourframework that do not engage in optimization and are partially or entirely unknown to the search engine. These models may use different LLM architectures, introduce new tasks, and/or use a new dataset for a known task. Using such a rich ecosystem, we aim at answering the following fundamental research questions that we believe are essential to understand the potential of developing \emph{search engines for machines}. Note that given the scope of these research questions and space limitation, we limit this study to reranking optimization, in which the search engine retrieves documents using BM25 and optimizes a personalized cross-encoder reranker based on the feedback received from the downstream RAG models.

\medskip

\noindent \textbf{RQ1: How does unified reranking perform compared to training individual rerankers for each RAG model?} Our experiments demonstrated that unified reranking performs either on par or significantly better than the same reranking model being trained for each downstream RAG model individually. In more detail, improvements are statistically significant for 61\% of downstream models and there is no statistically significant performance degradation across the 18 RAG models that engage in the training process. This critical finding  suggests that developing a unified search engine for machines is a promising research direction for the IR community to pursue further.

\medskip

\noindent \textbf{RQ2: How does unified reranking perform compared to training rerankers using the feedback obtained from all RAG systems that use the same dataset?} This research question helps us understand if unification across datasets and tasks can lead to downstream performance improvement. Our experiments suggest that again training a unified reranker performs on par or significantly better than the ones that are trained based on the feedback provided by all RAG models that use the same dataset. In more detail, $39\%$ of downstream RAG models in our experiments observe statistically significant improvement, while the majority observe no significant differences. These results are particularly important as they show knowledge transfer can occur across datasets. 

\medskip

\noindent \textbf{RQ3: How does ``personalizing'' search results for different RAG systems impact the performance?} We hypothesize that different RAG systems behave differently, thus ``personalizing'' search results for each RAG model can be beneficial. To achieve this, we feed a task and model identifier for each RAG model to our cross-encoder reranker, in addition to the query and document content. We observe that $50\%$ of downstream RAG models observe improvements by including task and model identifiers in relevance scoring. However, in the vast majority of cases, these improvements are not statistically significant. Therefore, we do not find personalization in our experimentation ecosystem substantially useful. We must acknowledge that  personalization often demonstrates its significance as the number of users increases and here we only have 18 users (i.e., the downstream RAG systems that engage in training). Therefore, this leaves several open questions for future work on exploring personalization in the context of \ourframework.

\medskip

\noindent \textbf{RQ4: How does unified reranking perform for new (unknown) RAG systems that use an observed (known) dataset?} Ideally, search engines should perform effectively and robustly for new users. This would demonstrate the generalizability of the retrieval model to unknown agents. To address this question, we introduce three retrieval-augmented generation models based on PEGASUS \cite{pegasus}, Mistral \cite{mistral}, and Llama2 \cite{llama2} to our experiments. These models did not engage in the optimization pipeline of \ourframework, thus are unknown to our unified reranker. We demonstrate that unified reranking significantly outperforms BM25 for all these new (unknown) RAG systems and performs comparably with a reranker that are trained on all other RAG systems that use the same dataset. 

\medskip

\noindent \textbf{RQ5: How does unified reranking perform for a known RAG system on a new (unknown) dataset that is similar to the ones used during training?} To address this research question, we use the open-domain variant of the SQuAD dataset \cite{squad} as a question answering data with short answers that is relatively similar to some of the datasets used during training. We observe that unified reranking leads to significant improvements compared to BM25 and since the dataset is new, other reranking alternatives are not applicable to this scenario.

\medskip

\noindent \textbf{RQ6: How does unified reranking perform for a known RAG system on entirely new tasks?} We take a step further and consider evaluation sets that are not closely related to our training datasets. For this purpose, we consider ELI5, an open-domain question answering dataset with long-form (passage-level) answers, as well as AY2 an entity linking dataset. Note that all datasets used during optimization target short text generation tasks. We observe that \ourframework outperforms BM25 in half of the cases and differences are not statistically significant. This suggests that reranking documents using the current implementation of \ourframework should only be employed when target downstream tasks are closely related to the ones used during training. 

\medskip

\noindent \textbf{RQ7: How does unified reranking perform for a new (unknown) RAG system on a new (unknown) dataset that is similar to the ones used during training?} We evaluate retrieval-augmented models based on PEGASUS, Mistral, and Llama2 on SQuAD for answering this question. None of these models were employed during training and as mentioned earlier, SQuAD is similar to some of the datasets used for training \ourframework. We observe that reranking results using \ourframework can lead to statistically significant improvements in this scenario. This emphasizes that the target task and data have more impact on the effectiveness of \ourframework, compared to the downstream RAG system that consumes the retrieval results.

\medskip

\noindent \textbf{RQ8: How does unified reranking perform for a new RAG system on an entirely new (unknown) task?} This is the most extreme case, where both the model architecture and the task are unknown to the search engine. We test PEGASUS, Mistral, and Llama2 on ELI5 and AY2 and observe that \ourframework struggles with improving BM25 in this scenario. We mark the development of robust and generalizable search engines for such scenarios as an important research direction for further exploration.

\medskip

\noindent \textbf{RQ9: How does \ourframework perform when different amount of training data is provided?} We plot the learning curve for all 18 RAG models that are involved in our training pipeline and observe that their performance often improves as more training data is provided. This finding suggests that as we introduce more data and more RAG systems to the \ourframework framework, we expect even further performance improvement. 

\medskip

We believe that these findings deepen our knowledge and understanding for developing effective search engines for downstream retrieval-augmented systems and smooth the path towards important developments in this area. We open-source the \ourframework codebase and release our trained model parameters.\footnote{\url{https://github.com/alirezasalemi7/uRAG}} Given the high cost of training and building such an ecosystem, we expect that this public release will substantially speed up research progress in this area.

\section{Related Work}

\subsubsection*{\textbf{Knowledge-Intensive Language Tasks}}

Unlike conventional NLP tasks like natural language understanding \cite{glue, superglue} and question answering \cite{mccann2019the}, where the task's input alone is sufficient for task completion, knowledge-intensive NLP tasks demand access to external knowledge sources to retrieve essential information for task execution. \citet{kilt} established KILT as a benchmark for knowledge-intensive NLP tasks. Encompassing a wide range of tasks such as open-domain question answering, fact checking, slot filling, and entity linking, KILT provides a comprehensive evaluation framework. The experiments conducted in this paper leverage datasets sourced from the KILT benchmark.

\subsubsection*{\textbf{Retrieval-Augmented Generation (RAG)}}

RAG \cite{rag} is a paradigm that integrates information retrieval (IR) and natural language generation (NLG). This approach aims to enhance the quality and relevance of generated content by leveraging external knowledge sources during the generation process \cite{asai2023selfrag, siriwardhana-etal-2023-improving}. Unlike traditional large language models that rely solely on knowledge learned during their pre-training, RAG systems dynamically fetch information from a knowledge source using a retriever, allowing them to generate contextually rich and factually accurate outputs \cite{reml}. RAG demonstrates versatility with numerous use cases and applications. It finds application in knowledge-grounding within textual \cite{kilt, rag, fid} or multi-modal \cite{dedr, murag, kat, salemi2023pretraining} problem domains. Additionally, RAG contributes to personalization efforts \cite{salemi2024lamp, salemi2024optimization} and addresses challenges such as reducing hallucination \cite{agrawal2023knowledge, shuster-etal-2021-retrieval-augmentation}.

A pivotal element in RAG systems is the retriever, tasked with obtaining the required information for the large language model to execute its task \cite{rag}. Typically, a either sparse (e.g., BM25 \cite{bm25}) or dense retrieval models (e.g., DPR \cite{dpr}, Contriever \cite{contriever}, ColBERTv2 \cite{santhanam-etal-2022-colbertv2}, etc.) is employed to efficiently retrieve information from knowledge sources. Subsequently, the large language model employs the retrieved information to accomplish its task. Notably, In-Prompt Augmentation (IPA) and Fusion-in-Decoder (FiD) \cite{fid} represent prominent approaches in this context. In IPA, the retrieved information is concatenated with the prompt, enabling the large language model to fulfill the task. On the other hand, FiD involves encoding each retrieved document and the prompt separately in the encoder of an encoder-decoder LLM, followed by concatenation in the decoder to generate a unified answer drawing from the information in all documents. For a more in-depth exploration of this approach, refer to the work by \citet{fid}.

\subsubsection*{\textbf{Optimizing Retriever in RAG pipelines}}

\begin{figure}
    \centering
    \includegraphics[width=\linewidth]{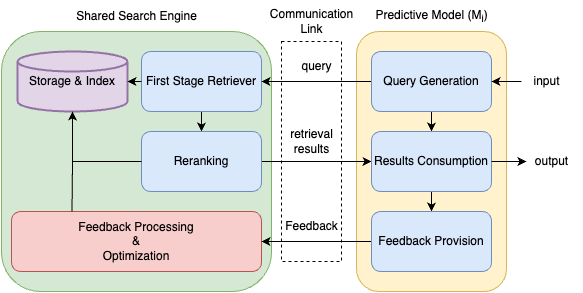}
    \caption{An overview of interactions between RAG models (also known as predictive models) and the unified search engine.}
    \label{fig:overview-framework}
\end{figure}

Joint training of a retriever model and a large language model has been explored in various studies. \citet{retriever-reader-distill} focuses on optimizing the retriever by leveraging the downstream performance of the LLM, using each individually retrieved document to guide the retriever towards retrieving documents with higher scores. Conversely, \citet{distil-fid} adopts a different strategy by utilizing cross-attention weights of the LLM instead of the downstream performance to distill knowledge from the LLM to the retriever. Additionally, \citet{sachan-etal-2021-end} introduces EMDR\textsuperscript{2}, an end-to-end approach that employs an expectation-maximization algorithm to optimize both the retriever and the LLM, maximizing the probability of generating the ground truth answer conditioned on the documents retrieved by the retriever. Moreover, \citet{atlas} presents a similar approach to \citet{distil-fid} and \citet{sachan-etal-2021-end}, utilizing the document's posterior distribution according to the LLM, conditioned on a single document, to perform distillation. \citet{rl-retriever-train} uses a multi-armed bandit algorithm to for the joint training of a retriever and reader. Most recently, \citet{StochasticRAG} proposes end-to-end RAG optimization through stochastic sampling without replacement, approximated by gumbel-top-k sampling.

Our work diverges from prior research in two key aspects. Firstly, we operate under the assumption that LLM parameters in downstream models are not accessible to the retrieval model. This stems from our conceptualization of the LLM as a user of the retrieval model, akin to how humans interact with search engines. In this analogy, search engines can only offer results to users and must rely on user feedback (e.g., clicks) to refine themselves, without altering how users engage with the results. Secondly, our focus extends beyond previous studies by considering scenarios where multiple LLMs use a shared search engine. Our objective is to optimize the search engine to enhance the collective performance of these diverse LLMs.


\section{The \ourframework Ecosystem}

\label{sec:urag-ecosystem}

Search engines are often designed for human users who submit unstructured queries, such as keyword queries and natural language questions. However, a paradigm shift is underway with the emergence of machine learning models \cite{reml}, especially large language models, possessing strong linguistic capabilities, memorization, and sometimes even reasoning to some extent \cite{zhao2023survey}. In the current paradigm known as REML \cite{reml}, machines, e.g., LLMs, engage in interactions with a retrieval model to acquire essential knowledge for executing their designated tasks. Most REML systems nowadays are in the form of retrieval-augmented text generation. Notably, with the introduction of real-world RAG systems, such as Bing Chat\footnote{\url{https://www.bing.com/chat}} and Google Bard,\footnote{\url{https://bard.google.com/}} humans now assume the role of users for these applications \cite{reml, salemi2024evaluating}, marking a transition from their previous role as users of web search engines. In this case, LLMs constitute the primary users of search engines. Exploring the design of a search engine capable of providing information to different LLMs based on their needs, each tailored for specific tasks, represents a valuable and promising research direction that this paper focuses on.

\paragraph{\textbf{\ourframework Formulation}}
Let $R_\theta$ denote a search engine parameterized by $\theta$ whose goal is to provide access to information from a corpus $C$ to a set of $n$ RAG models. The set of RAG models, that are essentially the users of $R_\theta$, is represented as $\mathbf{M}=\{M_1, ..., M_n\}$. For $R_\theta$, each $M_i$ is a black-box model and the search engine does not have access to the architecture, configuration, or parameters of RAG models. Each RAG model $M_i$ is designed to perform a specific task ${T_i}$ that requires access to information from the corpus $C$. These tasks are often called knowledge-intensive tasks. Each RAG model $M_i$ operates by taking the input $x$ and utilizing a communication link to interact with the search engine $R_\theta$. Through this interaction, the model produces an output denoted as $\hat{y}_{M_i}$, representing the result of the prediction process for the given input $x$. Formally, $\hat{y}_{M_i} = M_i(x; R_\theta)$. Inspired by the REML framework presented in \cite{reml}, Figure~\ref{fig:overview-framework} illustrates the interactions and components in \ourframework. 

\paragraph{\textbf{The Optimization Guideline and Process in \ourframework}} For each query submitted by the RAG model $M_i$, our goal is to develop a search engine that delivers a search result that benefits the downstream task conducted by  $M_i$. Thus, we assume that $M_i$s engage in an optimization process with $R_\theta$ with a pre-defined training guideline:
\begin{enumerate}

    \item Each RAG model $M_i$ must have a training set $\mathbf{D_i} = \{(x_1, y_1),$ $\cdots,(x_{|\mathbf{D_i}|}, y_{|\mathbf{D_i}|})\}$, where $(x_j, y_j)$ denotes the $j$\textsuperscript{th} input and expected output. There is no need to share the data with the search engine $R_\theta$.
    
    \item For each $(x, y) \in \mathbf{D}_i$, the RAG model $M_i$ must construct a query $q = \textsc{QGen}(x)$ to be consumed by the search engine $R_\theta$. The search engine returns a ranked list of documents in response. Let $\mathbf{Q}_i = \{\textsc{QGen}(x): \forall (x, y) \in \mathbf{D}_i\}$ be a set of all queries constructed from data points in $\mathbf{D}_i$.

    \item Each RAG model $M_i$ must identify a utility function $\textsc{Utility}_i$ that quantifies the end-to-end performance of the model in relation to its designated task ${T_i}$ and serves as a metric to measure the effectiveness of $M_i$ in performing its downstream task. $\textsc{Utility}_i$ can be any arbitrary function, differentiable or not, whose outputs will be provided as feedback to $R_\theta$ for optimization. For simplicity, this paper assumes that $\textsc{Utility}_i$ values are in the $[0, 1]$ range.
\end{enumerate}

According to the risk minimization framework, the primary goal of the search engine $R_\theta$ is to minimize the loss function defined based on the received utility values from downstream models:
\begin{equation}
    \label{eq:formulation}
    \theta^* = \operatorname*{arg\,min}_{{\theta}} \frac{1}{n} \sum_{i=1}^{n} \frac{1}{|\mathbf{Q}_i|}\sum_{q \in \mathbf{Q}_i} L(R_\theta(q; C), U_{iq})
\end{equation}
where $U_{iq}$ is computed by $M_i$ as follows:
\begin{equation}
    U_{iq} \coloneqq \textsc{Utility}_i(y, M_i(x; R_\theta(\textsc{QGen}_i(x); C)))
\end{equation}

It is important to note that various methods exist for aggregating utility across the users. For example, prioritizing utility enhancement for specific users may be more important than others. Alternatively, fairness or robustness could serve as criteria for aggregating the overall system utility. However, in this particular problem, our emphasis is on expected utility across inputs and users.

\section{System Implementation and Setup}

\begin{table}[t]
    \centering
    \caption{A list of datasets from KILT \cite{kilt} used in our experiments. The validation data from KILT is used as test sets. The first six datasets are used during the search engine training process. The last three datasets, on the other hand, are only introduced during inference to quantify generalizability of \ourframework. $^*$ Given the large training set in the original T-REx dataset, we only sampled 5\% of data for training our models.}
    \begin{tabular}{ll|ll}
        & \textbf{Dataset} & \textbf{\#train} & \textbf{\#test} \\ \hline
        \multicolumn{2}{l|}{\textbf{open-domain QA (short answer)}} & \\
        & Natural Questions & 87,372 & 2,837 \\
        & TriviaQA & 61,844 & 5,359 \\
        & HotPotQA & 88,869 & 5,600 \\\hdashline
        \multicolumn{2}{l|}{\textbf{fact verification}} & \\
        & FEVER & 104,966 & 10,444 \\\hdashline
        \multicolumn{2}{l|}{\textbf{slot-filling relation extraction}} & \\
        & zsRE & 147,909 & 3,724 \\
        & T-REx & 114,208$^*$ & 5,000 \\\hline
        \multicolumn{2}{l|}{\textbf{new data introduced during inference}} & \\
        & SQuAD & 87,599 & 10,570 \\
        & ELI5 & 272,634 & 1,507 \\
        & AY2 & 18,395 & 4,784 \\\hline
    \end{tabular}
    \label{tab:data}
\end{table}

\subsection{Training Data}

As we will discuss later, different tasks and datasets are used for training RAG models and providing feedback to $R_\theta$. A list of all datasets is presented in Table~\ref{tab:data}. Three diverse open-domain question answering datasets, i.e., Natural Questions (NQ) \cite{kwiatkowski-etal-2019-natural}, TriviaQA \cite{joshi-etal-2017-triviaqa}, and HotPotQA \cite{yang-etal-2018-hotpotqa}, are used in our experiments. The answers for all questions in these datasets are in the form of short text. HotPotQA focuses on questions that require multi-hop reasoning. One dataset is used for fact verification, called FEVER \cite{thorne-etal-2018-fever}, and finally, two slot-filling datasets for relation extraction are used in our experiments; zsRE \cite{levy-etal-2017-zero} and T-REx \cite{elsahar-etal-2018-rex}. All datasets are obtained from the KILT benchmark \cite{kilt}. 
Note that the last three datasets mentioned in Table~\ref{tab:data} are not used for optimizing $R_\theta$ and they are primarily employed for evaluating the generalizability of approaches.  

Note that since the T-REx training set includes approximately 2.2 million samples, we randomly select 5\% of them to train our models to speed up experiments. It is important to note that labels for test sets of these datasets are not publicly available. Therefore, as the public validation set was not used for training or hyperparameter tuning, we use the validation set directly to evaluate our models.

\begin{table}[t]
    \centering
    \caption{A list of RAG models used in our experiments for training and evaluation.}
    \vspace{-0.2cm}
    \resizebox{\linewidth}{!}{
    \begin{tabular}{l|llll}
         & \textbf{Task} & \textbf{Data} & \textbf{Utility Func.} & \textbf{LM} \\\hline
        $M_1$ & open-domain QA & NQ & Exact Match & RA-T5 \\
        $M_2$ & open-domain QA & NQ & Exact Match & RA-BART \\
        $M_3$ & open-domain QA & NQ & Exact Match & FiD \\\hdashline
        $M_4$ & open-domain QA & TriviaQA & Exact Match & RA-T5 \\
        $M_5$ & open-domain QA & TriviaQA & Exact Match & RA-BART \\
        $M_6$ & open-domain QA & TriviaQA & Exact Match & FiD \\\hdashline
        $M_7$ & open-domain QA & HotPotQA & Exact Match & RA-T5 \\
        $M_8$ & open-domain QA & HotPotQA & Exact Match & RA-BART \\
        $M_9$ & open-domain QA & HotPotQA & Exact Match & FiD \\\hline
        $M_{10}$ & fact verification & FEVER & Accuracy & RA-T5 \\
        $M_{11}$ & fact verification & FEVER & Accuracy & RA-BART \\
        $M_{12}$ & fact verification & FEVER & Accuracy & FiD \\\hline
        $M_{13}$ & slot filling & zsRE & Accuracy & RA-T5 \\
        $M_{14}$ & slot filling & zsRE & Accuracy & RA-BART \\
        $M_{15}$ & slot filling & zsRE & Accuracy & FiD \\\hdashline
        $M_{16}$ & slot filling & T-REx & Accuracy & RA-T5 \\
        $M_{17}$ & slot filling & T-REx & Accuracy & RA-BART \\
        $M_{18}$ & slot filling & T-REx & Accuracy & FiD \\\hline
    \end{tabular}
    }
    \label{tab:predictive_models}
\end{table}

\subsection{Downstream RAG Models for \ourframework Optimization}

\label{sec:predictive-models-imp}

To have a comprehensive study, we consider a total of 18 diverse RAG models during training in our experiments. Different RAG models conduct different tasks, are trained using different resources, are using different underlying models, and/or are consuming different number of retrieved documents.  Such diversity would enable us to evaluate the generalizability of different approaches. The downstream task conducted by models $M_1$ to $M_9$ is open-domain question answering. Models $M_{10}$, $M_{11}$, and $M_{12}$ deliver fact verification functionality to their users. Finally, models $M_{13}$ to $M_{18}$ perform slot filling for relation extraction. All RAG models are listed in \tablename~\ref{tab:predictive_models}. Each of these models is trained on a different dataset or uses a different large language model. We consider three retrieval-augmented large language models:
\begin{enumerate}[leftmargin=*]
    \item Retrieval-augmented T5 (RA-T5) is a language model based on T5-small \cite{t5} with 60 million parameters that consumes $k$ documents per input via in-prompt augmentation based on the following input format: ``\texttt{\{input\} context 1: \{doc1\} \ldots  context k: \{dock\}}'', where \texttt{\{input\}} is $x$ and \texttt{\{doci\}} denotes the content of the $i$\textsuperscript{th} retrieved document. Our T5 model has an input length limitation of 4096 tokens. Given this limitation, we feed $k=10$ documents to this model. Each RA-T5 model is fine-tuned on the corresponding training set. For example, the RA-T5 model in $M_1$ is trained on the Natural Questions dataset variant in KILT \cite{kilt} (see data statistics in Table~\ref{tab:data}).

    \item Retrieval-augmented BART (RA-BART) is BART-base \cite{bart} with 140 million parameters that uses the same augmentation approach for input format as RA-T5. BART goes under a different pre-training process than T5, thus, would exhibit a different behavior and performance. Additionally, it has an input length limitation of 1024 tokens. Therefore, we use only $k=4$ for augmenting BART using the results returned by $R_\theta$. Each RA-BART model is fine-tuned on the corresponding training set.

    \item Fusion-in-Decoder (FiD) \cite{fid} uses a different augmentation approach. Unlike RA-T5 and RA-BART that are based on in-prompt augmentation, FiD first encodes the input and each retrieved document separately and uses the concatenation of all document encodings as cross-attention for the decoder. FiD can thus only be done using encoder-decoder language models, for which we used T5-small \cite{t5} in our experiments. We use FiD-small in our experiments. The number of augmented documents is set to $k=10$ for FiD. Similarly, each FiD model is fine-tuned on the corresponding training set.
\end{enumerate}

All retrieval-augmented models listed in Table~\ref{tab:predictive_models} are fine-tuned separately. We use the AdamW optimizer with a weight decay of $10^{-2}$ and a learning rate of $5 \times 10^{-5}$ for $10$ epochs. Linear warmup is applied to the first $5\%$ of training steps. The effective batch size is set to 64, achieved through gradient accumulation. Each model is trained using different computation resources, including up to 8 A100, 1080ti, and 2080ti Nvidia GPUs. We employ BM25\cite{bm25}, implemented in Pyserini \cite{10.1145/3404835.3463238}, to retrieve documents for the training of the aforementioned models. To train the models, a seq2seq loss (i.e., cross-entropy between the predicted token's probability and the ground truth token distribution) is employed. 

As detailed in Section~\ref{sec:urag-ecosystem}, each RAG model is required to define a utility function that assesses its performance. To select the utility function for our RAG models, we adhere to the recommended metrics outlined by the KILT benchmark \cite{kilt}. Thus, we choose the evaluation metric of each dataset as the utility function for the models that perform their task on that dataset. Table~\ref{tab:predictive_models} illustrates the utility functions assigned to the RAG models utilized in this paper. In general, we employ Exact Match (EM) for short-form question answering datasets, ROUGE-L for long-form question answering datasets, and accuracy as the utility function for the remaining tasks. For Exact Match, we follow the post-processing approaches introduced by \citet{squad}.

\subsection{Search Engine Implementation in \ourframework}

We adopt a two-stage cascaded retrieval pipeline for implementing $R_\theta$. A wikipedia dump provided by the KILT benchmark is used as the unstructured knowledge source.\footnote{The retrieval corpus is available at \url{https://dl.fbaipublicfiles.com/ur/wikipedia_split/psgs_w100.tsv.gz}} We adhere to the pre-processing method outlined by \citet{dpr}, where each document is divided into passages, each with a maximum length of 100 words. Additionally, we concatenate the document title with each passage to form the documents in the retrieval corpus. This corpus consists of approximately 36 million passages.

We use Pyserini's \cite{10.1145/3404835.3463238} implementation of BM25 with default parameters for implementing the first stage that takes the query string $q$ and retrieves 100 documents. The second stage model is a cross-encoder re-ranker that takes the query and document text as input and produces a scalar as the relevance score. To ``personalize'' the result of the retrieval model, we contextualize the cross-encoder representation based on $M_i$ and its downstream task $T_i$. Following \citet{bert-reranker}, we use a linear projection  over the representation associated with the start token (i.e., \texttt{[CLS]}) to obtain the relevance score, as follows:
\begin{equation}
    s_{d} = \textsc{Encoder}_{\text{[CLS]}}(\texttt{tid}; \texttt{mid}; q; d) \cdot W
\end{equation}
where \texttt{tid} and \texttt{mid} represent the downstream task and model identifiers, respectively. The sign $;$ denotes concatenation with a separation token. $W \in \mathbb{R}^{D \times 1}$ is a linear projection layer, where $D$ represents the embedding dimensionality of the encoder. In this paper, we utilize the BERT-base~\cite{bert} as the text encoder. Once all the 100 documents are re-ranked, the top $k$ documents are selected as the model's output, denoted as $L_q$. Here, the search engine logs the query and the returned results for later use, e.g., using them for optimization when the feedback from the RAG model is received. 


For optimization, $R_\theta$ solicits feedback one by one for each document $d \in L_q$. We then apply a threshold $\tau_{(\text{tid},\text{mid})}$ to the feedback $U_i$ received for each document; the feedback higher than or equal to the threshold is considered as a positive feedback, negative otherwise. Based on these feedbacks, we create a set of positive ($D^q_{\text{pos}}$) and negative ($D^q_{\text{neg}}$) documents for each query. We then apply a binary cross-entropy loss function to train the cross-encoder model:

\begin{equation}
    L = - \frac{1}{|Q|} \sum_{q \in Q} (\sum_{d \in D^q_{\text{pos}}} \log \sigma(s_d) + \sum_{d \in D^q_{\text{neg}}} \log \sigma(1 - s_d)) 
\end{equation}

\noindent
where $Q$ is the set of all queries used for training. Note that we randomly substitute the model ID and task ID with the \textit{``unk''} token for $10\%$ of training samples. This aids the model in adapting to new tasks and models during inference. The rationale behind this approach is to train the model to correctly assign labels to samples even when it lacks information about the specific task and model being performed. For training $R_\theta$, we use Adam optimizer \cite{adam} with a learning rate of $10^{-5}$ for two epochs. Linear warmup is applied to the first $5\%$ of training steps. The effective batch size is set to 512, which is achieved through gradient accumulation. Values of $k = 32$ and $\tau_{(\text{tid},\text{mid})} = 0.5$ are consistently employed in all experiments. The maximum input length for this model is set to 256 tokens.




\section{Empirical Results and Findings}

\begin{table*}[t]
    \centering
    \caption{Downstream performance obtained by 18 RAG models listed in Table~\ref{tab:predictive_models} consuming different retrieval results. The overal performance is the macro-average of the performance of 18 RAG models. Superscripts $1$, $2$, $3$, $4$, and $5$ denote statistically significant improvements in the performance compared to BM25, Contriever, $\text{Reranker}_{\text{individual}}$, $\text{Reranker}_{\text{dataset}}$, and  $\text{Reranker}_{\text{unified}}$ w/o personalization, respectively. We used McNemar statistical test for significant test ($p<0.05$).}
    \vspace{-0.2cm}
    \resizebox{\linewidth}{!}{
    \begin{tabular}{ll|cc|ll|ll}
        \textbf{RAG} & \multirow{2}{*}{\textbf{Data \& Metric}} & \multirow{2}{*}{\textbf{BM25}} & \multirow{2}{*}{\textbf{Contriever}} & \multirow{2}{*}{\textbf{$\text{Reranker}_{\text{individual}}$}}& \multirow{2}{*}{\textbf{$\text{Reranker}_{\text{dataset}}$}} & \textbf{$\text{Reranker}_{\text{unified}}$} & \multirow{2}{*}{\textbf{$\text{Reranker}_{\text{unified}}$}} \\
        \textbf{Model} & & & & & & \textbf{w/o personalization} \\\hline
        $M_1$ & NQ - EM & 28.05 & 22.55 & 36.76 & 37.39 & 37.18 & 37.82$^{12}$ \\
        $M_2$ & NQ - EM & 33.09 & 23.68 & 40.07 & 40.50 & 42.29 & 42.12$^{1234}$ \\
        $M_3$ & NQ - EM & 29.64 & 23.69 & 40.50 & 41.14 & 40.47 & 42.37$^{12345}$ \\\hdashline
        $M_4$ & TriviaQA - EM & 51.35 & 44.33 & 59.28 & 60.25 & 60.30 & 60.68$^{123}$ \\
        $M_5$ & TriviaQA - EM & 57.52 & 48.49 & 64.76 & 67.23 & 67.30 & 68.12$^{123}$ \\
        $M_6$ & TriviaQA - EM & 60.48 & 49.30 & 67.44 & 68.63 & 69.14 & 68.74$^{123}$ \\\hdashline
        $M_7$ & HotPotQA - EM & 27.51 & 18.80 & 29.92 & 30.91 & 31.32 & 31.33$^{123}$ \\
        $M_8$ & HotPotQA - EM & 31.21 & 20.78 & 35.03 & 34.62 & 34.10 & 34.85$^{124}$ \\
        $M_9$ & HotPotQA - EM & 29.48 & 20.43 & 32.54 & 32.71 & 32.96 & 33.46$^{1234}$ \\\hline
        $M_{10}$ & FEVER - Accuracy & 86.83 & 84.21 & 86.24 & 86.83 & 87.02 & 86.46$^{2}$ \\
        $M_{11}$ & FEVER - Accuracy & 87.54 & 84.37 & 84.38 & 87.54 & 86.78 & 85.99$^{23}$ \\
        $M_{12}$ & FEVER - Accuracy & 87.04 & 86.74 & 86.02 & 87.04 & 87.21 & 86.55$^{23}$ \\\hline
        $M_{13}$ & zsRE - Accuracy & 55.37 & 38.77 & 60.39 & 59.98 & 61.41 & 61.09$^{124}$ \\
        $M_{14}$ & zsRE - Accuracy & 51.42 & 29.05 & 59.29 & 58.96 & 60.92 & 60.58$^{1234}$ \\
        $M_{15}$ & zsRE - Accuracy & 55.42 & 37.35 & 60.47 & 60.66 & 62.08 & 62.13$^{1234}$ \\\hdashline
        $M_{16}$ & T-REx - Accuracy & 70.88 & 56.94 & 73.58 & 72.86 & 73.70 & 72.92$^{12}$ \\
        $M_{17}$ & T-REx - Accuracy & 75.16 & 58.30 & 80.04 & 80.18 & 79.64 & 79.94$^{12}$ \\
        $M_{18}$ & T-REx - Accuracy & 78.88 & 65.06 & 80.78 & 80.34 & 80.86 & 80.24$^{12}$ \\\hline
        Overall & -- & 55.38 & 45.15 & 59.86 & 60.43 & 60.81 & 60.85$^{1234}$\\\hline
    \end{tabular}
    }
    \label{tab:main-results}
\end{table*}



This section is dedicated to presenting supporting results that address the research questions outlined in Section~\ref{sec:intro}.

\subsubsection*{\textbf{RQ1: How does unified reranking perform compared to training individual rerankers for each RAG model?}}

The results presented in Table~\ref{tab:main-results} offer insights into answering this research question. It is evident all reranking models applied to BM25 lead to improved downstream performance for all 18 models we have in our experiments. The only exception is the models trained and evaluated on FEVER dataset for fact verification. As a point of reference, we also compare the results with Contriever \cite{contriever}--an effective zero-shot dense retrieval models. Contriever consistently underperforms BM25 for all downstream models $M_1$ to $M_{18}$. 

Additionally, comparing the results obtained by $\text{Reranker}_{\text{unified}}$ and $\text{Reranker}_{\text{individual}}$ in Table~\ref{tab:main-results} shows that developing a unified retrieval models for all retrieval-augmented models achieves a better performance for 78\% of models (i.e., 14 out of 18 models; all except $M_8$ on HotPotQA and $M_{16}$ to $M_{18}$ on T-REx). For the four models that unified retrieval does not show improvement, we do not observe statistically significant differences between model performances. Thus, we can conclude unified retrieval performs on par or significantly better than individual retrieval optimization in almost all cases. Statistically significant improvement has been observed in 11 out of 18 downstream models. Looking at the overall performance, we show that unified retriever leads to statistically significantly higher performance compared to individual retrievers. 



In conclusion, the discussion above suggests that our approach, which leverages the feedback from all RAG models across all tasks to optimize the search engine, is significantly superior to the alternative of training a single retrieval model for each RAG model, which is the current standard in developing RAG systems.

\subsubsection*{\textbf{RQ2: How does unified reranking perform compared to training rerankers using the feedback obtained from all RAG systems that use the same dataset?}}

Instead of retrieval optimization for individual models, one may suggest to train a retrieval model for the downstream models that are using the same task and dataset, which we call $\text{Reranker}_{\text{dataset}}$. Comparing the results obtained by this approach and $\text{Reranker}_{\text{unified}}$ in Table~\ref{tab:main-results} suggest that training a unified reranker across the feedback of all users on average statistically significantly outperforms the reranker trained on the feedbacks of models performing on the same dataset. In more details, unified optimization performs better for $72\%$ of models (i.e., 13 out of 18 models; all models except $M_{10}$, $M_{11}$, $M_{12}$,  $M_{17}$, and $M_{18}$, that are all performing on either FEVER or T-REx datasets). Note that there is statistically significant differences in $M_{11}$ and $M_{12}$ users. However, in other cases, there is not a statistically significant difference between the results for $M_{10}$,  $M_{17}$, and $M_{18}$. Considering overall performance across models, we observe statistically significant improvements when using unified training. 


In summary, the observations above indicate that for all downstream RAG models, leveraging feedback from all RAG models across tasks for optimizing the reranking model is either on par or significantly superior to the alternative approach of training a  retrieval model for each dataset. Therefore, retrieval models can benefit from knowledge transfer across datasets and tasks for RAG.

\subsubsection*{\textbf{RQ3: How does ``personalizing'' search results for different RAG systems impact the performance?}}

To assess the impact of personalization on the search engine for each RAG model, an experiment was conducted where a reranker was trained similarly to unified reranker but without incorporating task and model IDs as the input of reranker. The results of this experiment are illustrated in Table~\ref{tab:main-results}. The results suggest that personalization leads to an improvement on average across all users, but it is not significant.

More specifically, the improvements resulting from personalization are most pronounced in the question answering tasks. This is attributed to the similarity among most questions in these datasets, making it challenging for the non-personalized model to discern the specific task associated with a given query. Hence, personalization aids the reranker in identifying the task for which the input is intended and subsequently reranking the documents accordingly. Conversely, the improvements in the slot filling tasks (i.e., T-REx and zsRE) are smaller and for fewer users. Additionally, personalization adversely affects performance in the FEVER dataset. These observations indicate that while, on average, our approach for personalization was effective, there is potential for a further study on how to effectively personalize the search engine for RAG models, which is an important topic for future exploration.

\begin{table}[t]
    \centering
    \caption{A list of RAG models used in our experiments for evaluating generalizability of \ourframework. ELI5 is also an open-domain QA dataset, but the expected outputs are longer (e.g., sentences or a paragraph) compared to other datasets.}
    \vspace{-0.2cm}
    \resizebox{\linewidth}{!}{
    \begin{tabular}{l|llll}
         & \textbf{Task} & \textbf{Data} & \textbf{Utility Func.} & \textbf{LM} \\\hline
        $M_{19}$ & entity linking & AY2 & Accuracy & RA-T5 \\
        $M_{20}$ & long-form QA & ELI5 & ROUGE-L & RA-T5 \\
        $M_{21}$ & open-domain QA & SQuAD & Exact Match & RA-T5 \\
        $M_{22}$ & entity linking & AY2 & Accuracy & RA-BART \\
        $M_{23}$ & long-form QA & ELI5 & ROUGE-L & RA-BART \\
        $M_{24}$ & open-domain QA & SQuAD & Exact Match & RA-BART \\
        $M_{25}$ & open-domain QA & NQ & Exact Match & RA-PEGASUS \\
        $M_{26}$ & open-domain QA & TriviaQA & Exact Match & RA-PEGASUS \\
        $M_{27}$ & long-form QA & ELI5 & ROUGE-L & RA-PEGASUS \\
        $M_{28}$ & open-domain QA & SQuAD & Exact Match & RA-PEGASUS \\
        $M_{29}$ & open-domain QA & NQ & Exact Match & RA-Mistral \\
        $M_{30}$ & open-domain QA & TriviaQA & Exact Match & RA-Mistral \\
        $M_{31}$ & long-form QA & ELI5 & ROUGE-L & RA-Mistral \\
        $M_{32}$ & open-domain QA & SQuAD & Exact Match & RA-Mistral \\
        $M_{33}$ & open-domain QA & NQ & Exact Match & RA-Llama2 \\
        $M_{34}$ & open-domain QA & TriviaQA & Exact Match & RA-Llama2 \\
        $M_{35}$ & long-form QA & ELI5 & ROUGE-L & RA-Llama2 \\
        $M_{36}$ & open-domain QA & SQuAD & Exact Match & RA-Llama2 \\\hline
    \end{tabular}
    }
    \vspace{-0.3cm}
    \label{tab:new_predictive_models}
\end{table}

\begin{table}[t]
    \centering
    \caption{The performance on downstream task obtained for new (unknown) models applied to the datasets known to $R_\theta$. Superscript $^1$ and $^2$ denote statistically difference between our model and BM25 and {$\text{Reranker}_{\text{dataset}}$}, respectively ($p<0.05$).}
    \vspace{-0.3cm}
    \resizebox{\linewidth}{!}{
    \begin{tabular}{ll|cc|c}
        \textbf{RAG} & \multirow{2}{*}{\textbf{Data \& Metric}} & \multirow{2}{*}{\textbf{BM25}} & \multirow{2}{*}{\textbf{$\text{Reranker}_{\text{dataset}}$}} &  \multirow{2}{*}{\textbf{$\text{Reranker}_{\text{unified}}$}} \\
        \textbf{Model} & & &  \\\hline
        $M_{25}$ & NQ - EM & 33.38 & 45.08 & \textbf{45.29$^{1}$} \\
        $M_{29}$ & NQ - EM & 15.15 & \textbf{21.36} & 21.25$^{1}$ \\
        $M_{33}$ & NQ - EM & 11.13 & 16.53 &  \textbf{16.56$^{1}$} \\\hdashline
        $M_{26}$ & TQA - EM & 53.53 & 61.74 & \textbf{62.39$^{1}$} \\
        $M_{30}$ & TQA - EM & 44.74 & \textbf{50.40} & 50.19$^{1}$ \\
        $M_{34}$ & TQA - EM & 25.86 & \textbf{29.63} & 29.22$^{1}$ \\\hline
    \end{tabular}
    }
    \vspace{-0.4cm}
    \label{tab:new_model_old_data}
\end{table}


\subsubsection*{\textbf{RQ4: How does unified reranking perform for new (unknown) RAG systems that use an observed (known) dataset?}}
We design an experiment in which we incorporate three new RAG models with  different architecture and pre-training process:
\begin{enumerate}
    \item Retrieval-augmented PEGASUS (RA-PEGASUS) uses in-prompt augmentation with the PEGASUS language model \cite{pegasus} which has 570 million parameters. We use the same input format as used by RA-T5 and RA-BART (see Section~\ref{sec:predictive-models-imp}). PEGASUS has a 1024 maximum token limit as its input; thus we limit this model to consume $k=4$ retrieved documents. RA-PEGASUS models are also fined-tuned on the training set of the corresponding datasets. The fine-tuning details are the same as other retrieval-augmented models presented in Section~\ref{sec:predictive-models-imp}. 

    \item Retrieval-augmented Mistral (RA-Mistral) also uses in-prompt augmentation for Mistral LLM with 7 billion parameters \cite{mistral}, with no fine-tuning. Instead, to increase the diversity of models in our experiments, we use few-shot in-context learning for this approach, where five training examples are presented to the model in its prompt. Mistral can accept up to 32k tokens as input (including the few-shot examples). We feed $k=5$ retrieved documents to this model.\footnote{RA-Mistral uses the following prompt: \texttt{<s>[INST] you are a helpful assistant. following the answer patterns in the provided examples, answer the question concisely using the hints without any explanation. \{$\text{shot}_1$\} ... \{$\text{shot}_m$\} [/INST] question: \{input\}? hint: \{$\text{context}_1$\} ... \{$\text{context}_k$\} answer: }}

    \item Retrieval-augmented Llama2 (RA-Llama2) uses an in-prompt augmentation approach similar to RA-Mistral, but using Llama2 with 13 billion parameters\cite{llama2}. It also uses few-shot in-content learning, but with two examples instead (due to the input length limitation). Llama2 can accept up to 4096 tokens as input. RA-Llama2 uses $k=3$ retrieved documents.\footnote{RA-Llama2 uses the following prompt: \texttt{{you are a helpful assistant. following the answer patterns in the provided examples, answer the question concisely using the hints without any explanation. \{$\text{shot}_1$\} ... \{$\text{shot}_m$\} question: \{input\}? hint: \{$\text{context}_1$\} ... \{$\text{context}_k$\} answer: }}} 
\end{enumerate}

To address this research question, we evaluate these models on two existing datasets from our pipeline: Natural Questions (NQ) and TriviaQA (TQA). These models are denoted as $M_{25}$, $M_{26}$, $M_{29}$, $M_{30}$, $M_{33}$, and $M_{34}$, as is presented in Table~\ref{tab:new_predictive_models}. Since these models are new to our rerankers, their model and task ID is set to \texttt{unknown}, thus no personalization is conducted for these models. 

By comparing our unified reranking model with BM25 without reranking in Table~\ref{tab:new_model_old_data}, we observe significant improvements in all cases in comparison with BM25. The improvements range from 11\% to about $50\%$. Larger improvements are observed for the models evaluated on the Natural Questions dataset, which was also the case for the models used in our training process (see Table~\ref{tab:main-results}). Moreover, an alternative approach involves leveraging $\text{Reranker}_{\text{dataset}}$, which has been trained using the feedbacks from multiple large language models. The results of this alternative approach are also presented in Table~\ref{tab:new_model_old_data}. The results reveal that, for certain users, $\text{Reranker}_{\text{unified}}$ yields superior results, while for others, $\text{Reranker}_{\text{dataset}}$ performs better. Additionally, there is no statistically significant difference in performance between these models, suggesting comparable effectiveness in leveraging feedback from multiple large language models when it comes to serving a new large language model. 

\begin{table}[t]
    \centering
    \caption{The performance for new (unknown) models applied to the new datasets. SQuAD is a question answering dataset with short answers, thus similar to the datasets used during training such as NQ and TriviaQA. ELI5 focuses on long-form question answering, a task fundamentally different from datasets used during $R_\theta$ training. AY2 is a dataset for entity linking, which is not used for training $R_\theta$. Superscript $^*$ denotes significant improvement over BM25 ($p < 0.05$).}
    \vspace{-0.3cm}
    \begin{tabular}{ll|c|c}
        \textbf{RAG} & \multirow{2}{*}{\textbf{Data \& Metric}} & \multirow{2}{*}{\textbf{BM25}} & \multirow{2}{*}{\textbf{$\text{Reranker}_{\text{unified}}$}} \\
        \textbf{Model} & & &  \\\hline
        $M_{21}$ & SQuAD - EM & 35.91 & \textbf{41.19$^*$} \\
        $M_{24}$ & SQuAD - EM & 32.92 & \textbf{36.70$^*$} \\\hdashline
        $M_{19}$ & AY2 - Accuracy & \textbf{75.52} &  73.3 \\
        $M_{22}$ & AY2 - Accuracy & 81.98 & \textbf{82.08} \\\hdashline
        $M_{20}$ & ELI5 - ROUGE-L & 15.40 & \textbf{15.42} \\
        $M_{23}$ & ELI5 - ROUGE-L & \textbf{23.42} & 23.24 \\\hline
    \end{tabular}
    \label{tab:old_model_new_data}
\end{table}

\begin{table}[t]
    \centering
    \caption{The performance for new (unknown) models applied to new datasets that are unknown to $R_\theta$. Superscript $^*$ denotes statistically significant improvement over BM25 ($p < 0.05$).}
    \vspace{-0.2cm}
    \begin{tabular}{ll|c|c}
        \textbf{RAG} & \multirow{2}{*}{\textbf{Data \& Metric}} & \multirow{2}{*}{\textbf{BM25}} & \multirow{2}{*}{\textbf{$\text{Reranker}_{\text{unified}}$}} \\
        \textbf{Model} & & &  \\\hline
        $M_{28}$ & SQuAD - EM & 36.37 & \textbf{42.04$^*$} \\
        $M_{32}$ & SQuAD - EM & 13.01 & \textbf{15.18$^*$} \\
        $M_{36}$ & SQuAD - EM & 10.31 &  \textbf{12.06$^*$} \\\hdashline
        $M_{27}$ & ELI5 - ROUGE-L & \textbf{19.84} & 19.79 \\
        $M_{31}$ & ELI5 - ROUGE-L & \textbf{22.03} & 21.98 \\
        $M_{35}$ & ELI5 - ROUGE-L & \textbf{19.16} & 18.77 \\\hline
    \end{tabular}
    \label{tab:new_model_new_data}
    \vspace{-0.3cm}
\end{table}




\begin{figure*}
    \centering
    \includegraphics[width=\textwidth]{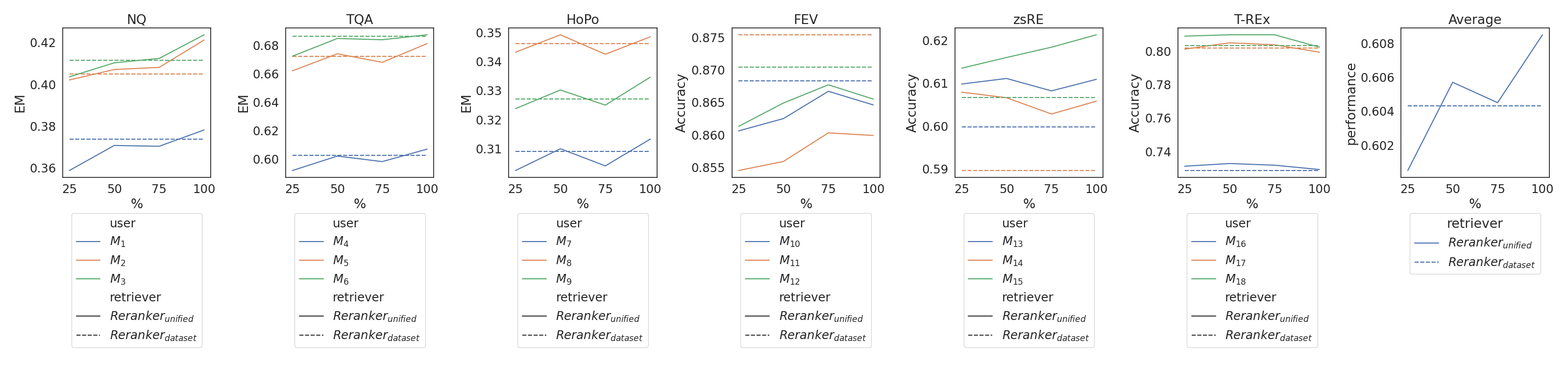}
    \vspace{-0.8cm}
    \caption{The performance of unified retrieval model using different percentages of training data. The dashed line indicates that the model is trained on the full dataset. The far right plot demonstrate the overall average performance across all datasets.}
    \label{fig:amount-data-performance}
\end{figure*}

\subsubsection*{\textbf{RQ5: How does unified reranking perform for a known RAG system on a new (unknown) dataset that is similar to the ones used during training?}}

To study this question, we conduct an experiment in which RA-BART and RA-T5 are employed as known models to the search engine for a new dataset, SQuAD \cite{squad}, which performs a known task (i.e., short-form open-domain question answering). These models are encoded as $M_{21}$ and $M_{24}$ (see Table~\ref{tab:new_predictive_models}). We use the open-domain variant of SQuAD \cite{squad} where the gold passage associated to each question is unknown.  The results of this experiment are presented in Table~\ref{tab:old_model_new_data}. The results suggest that when the new task is similar to the known tasks for the search engine, as in the case of SQuAD being similar to short-form question answering datasets used during training (i.e., NQ, TQA), significant improvements can still be achieved in comparison with the baseline (i.e., BM25). Note that the selection of baselines in this experiment is influenced by the consideration that other reranker baselines listed in Table~\ref{tab:main-results} are customized for specific model or dataset; they are not applicable to new models and datasets. 

\subsubsection*{\textbf{RQ6: How does unified reranking perform for a known RAG system on entirely new tasks?}}

To study this question, we conduct an experiment in which RA-BART and RA-T5 are employed as known models to the search engine for new tasks. ELI5 \cite{fan-etal-2019-eli5} serves as a long-form question answering task and AY2 \cite{hoffart-etal-2011-robust} as an entity linking dataset, which are new tasks for the search engine.  These models are denoted as $M_{19}$, $M_{20}$, $M_{22}$, and $M_{23}$ (see Table~\ref{tab:new_predictive_models}). The results of this experiment, presented in Table~\ref{tab:old_model_new_data}, suggest that when tasks are considerably different from the ones observed during training, unified retrieval as implemented in this approach is not likely to help. Indeed, unified retrieval model demonstrates superior performance for some users; however, there is no statistical difference between the unified retrieval model and the baseline. This creates an important opportunity for future work to focus on generalizability and adaptability of \ourframework to unknown tasks.


\subsubsection*{\textbf{RQ7: How does unified reranking perform for a new (unknown) RAG system on a new (unknown) dataset that is similar to the ones used during training?}}

To answer this question, in this experiment, we utilize RA-PEGASUS, RA-Mistral, and RA-Llama2 on SQuAD. These are models that did not exist at the time of training $R_\theta$ and SQuAD is a new (unknown) dataset for $R_\theta$ but related to NQ and TriviaQA. These models are encoded as $M_{28}$, $M_{32}$, and $M_{36}$ (see Table~\ref{tab:new_predictive_models}). The results of this experiment are displayed in Table~\ref{tab:new_model_new_data}. The findings indicate that when the task is similar to those encountered during model training (e.g., SQuAD in this case similar to NQ and TriviaQA in the training tasks), significant improvements over the baseline can be achieved even though the RAG model is unknown to the search engine.

\subsubsection*{\textbf{RQ8: How does unified reranking perform for a new RAG system on an entirely new (unknown) task?}}

In the most extreme case, both the task and the model are unknown to the search engine. Here, we utilize RA-PEGASUS, RA-Mistral, and RA-Llama2 on ELI5 dataset as a new task that is relatively different from the tasks used during training. These models are encoded as $M_{27}$, $M_{31}$, and $M_{35}$. The results in Table~\ref{tab:new_model_new_data} indicate that when the task is substantially different from those available during training, as is in ELI5 (i.e., long-form question answering), there is a drop in performance, though not statistically significant.

\subsubsection*{\textbf{RQ9: How does \ourframework perform when different amount of training data is provided?} }

We vary the amount of training data obtained from each downstream RAG model to $25\%$, $50\%$, $75\%$, and $100\%$ of the data used in other experiments. The results in  Figure~\ref{fig:amount-data-performance} show that even using $25\%$ of the training data, our approach is sufficiently effective to outperform the baseline trained on the full datasets (i.e., $\text{Reranker}_{\text{dataset}}$) in average. This trend continues when our approach consumes more data where it outperforms the baseline significantly when trained on the full datasets.

Furthermore, for most users, augmenting the training data improves the performance. Particularly noteworthy is the fact that for the FEVER and T-REx datasets, the optimal performance is achieved when utilizing 75\% of the training data. Importantly, for the majority of users, excluding $M_1$, $M_6$, $M_{10}$, $M_{11}$, and $M_{12}$, our approach either outperforms or performs equivalently to the baseline, trained on the full datasets, by observing 50\% of the training data. This underscores the effectiveness of unified reranking, especially when confronted with limited training data, for most users.

\section{Conclusion \& Future Work}
We studied the promises and challenges in developing a search engine for multiple downstream RAG systems. We introduced the \ourframework ecosystem that enabled us to conduct large-scale experimentation to answer some fundamental research questions in this area. We showed that optimizing a unified reranker leads to significant improvements compared to the current standard recipe--i.e., learning a retrieval model per downstream RAG model. We highlighted the improvements observed by the unified reranker is generalizable to new (unknown) models that are based on known datasets or unknown datasets that are closely related to the ones used in the training procedure. We further explained the challenges in personalizing rerankers for downstream RAG models and generalizing to unknown downstream tasks.

This paper opens up a wide range of future research directions. Through the proposed \ourframework ecosystem, future work can focus on (1) optimal aggregation of feedback from various downstream RAG models, (2) calibrating their feedback, (3) considering multiple utility functions per downstream model, (4) expanding the work to retrieval model optimization instead of reranking, (5) studying novel regularization techniques for improving \ourframework generalization, (6) studying online optimization of methods for \ourframework as downstream RAG models interact with the retrieval model, (7) exploring interleaving and counterfactual learning approaches in the context of \ourframework, (8) going beyond text generation and extending to a more general REML scenario, and many more.

\section*{Acknowledgments}
The authors would like to thank Fernando Diaz for invaluable discussions. This work was supported in part by the Center for Intelligent Information Retrieval, in part by NSF grant number 2143434, in part by the Office of Naval Research contract number N000142212688, in part by Lowe’s, and in part by an Amazon Research Award, Fall 2022 CFP. Any opinions, findings and conclusions or recommendations expressed in this material are those of the authors and do not necessarily reflect those of the sponsor. 

\bibliographystyle{ACM-Reference-Format}
\balance
\bibliography{XX-references}

\end{document}